\documentclass{article}
\usepackage{subcaption}
\usepackage{microtype}
\usepackage{graphicx}
\usepackage{booktabs}
\usepackage{listings}
\usepackage{wrapfig}
\usepackage{colortbl}
\usepackage{multirow}
\usepackage{subcaption}
\usepackage{placeins}
\usepackage{hyperref}

\usepackage{PRIMEarxiv}
\usepackage{amsmath}
\usepackage{amssymb}
\usepackage{mathtools}
\usepackage{amsthm}
\usepackage{natbib}
\usepackage[table,xcdraw]{xcolor}
\usepackage{pifont}
\usepackage[capitalize,noabbrev]{cleveref}
\usepackage{xcolor}
\usepackage{makecell}
\usepackage{xfrac} 
\usepackage{hyperref}
\definecolor{darkblue}{RGB}{8,54,104}
\definecolor{darkred}{RGB}{150,14,39}
\definecolor{nvidia-green}{RGB}{118, 185, 0}
\definecolor{blue}{RGB}{52,152,219}
\definecolor{hypernova-purple}{RGB}{142, 68, 173}
\theoremstyle{plain}

\theoremstyle{definition}

\theoremstyle{remark}

\usepackage[textsize=tiny]{todonotes}
\usepackage{authblk}
\title{Extending Puzzle for Mixture-of-Experts Reasoning Models with Application to GPT-OSS Acceleration
\\ \large{Technical Report}}

\author{Akhiad Bercovich}
\author{Nir Ailon}
\author{Vladimir Anisimov}
\author{Tomer Asida}
\author{Nave Assaf}
\author{Mohammad Dabbah}
\author{Ido Galil}
\author{Amnon Geifman}
\author{Yonatan Geifman}
\author{Izhak Golan}
\author{Roi Koren}
\author{Itay Levy}
\author{Zach Moshe}
\author{Pavlo Molchanov}
\author{Najeeb Nabwani}
\author{Mostofa Patwary}
\author{Omri Puny}
\author{Tomer Ronen}
\author{Itamar Schen}
\author{Elad Segal}
\author{Ido Shahaf}
\author{Oren Tropp}
\author{Ran Zilberstein}
\author{Ran El-Yaniv}

\affil{NVIDIA, \texttt{\{abercovich, relyaniv\}@nvidia.com}}

\begin{document}

\begin{figure}[t!]
    \centering
    \includegraphics[width=100pt]{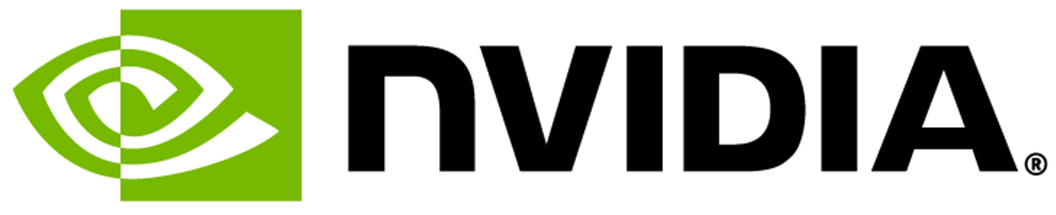}
\end{figure}

\maketitle

\begin{abstract}
Reasoning-focused LLMs improve answer quality by generating longer reasoning traces, but the additional tokens dramatically increase serving cost, motivating inference optimization.
We extend and apply Puzzle, a post-training neural architecture search (NAS) framework, to gpt-oss-120B to produce \emph{gpt-oss-puzzle-88B}, a deployment-optimized derivative.
Our approach combines heterogeneous MoE expert pruning, selective replacement of full-context attention with window attention, FP8 KV-cache quantization with calibrated scales, and post-training reinforcement learning to recover accuracy, while maintaining low generation length.
In terms of per-token speeds, on an 8$\times$H100 node we achieve 1.63$\times$ and 1.22$\times$ throughput speedups in long-context and short-context settings, respectively.
gpt-oss-puzzle-88B also delivers throughput speedups of 2.82$\times$ on a single NVIDIA H100 GPU.
However, because token counts can change with reasoning effort and model variants, per-token throughput (tok/s) and latency (ms/token) do not necessarily lead to end-to-end speedups: a 2$\times$ throughput gain is erased if traces grow 2$\times$.
Conversely, throughput gains can be spent on more reasoning tokens to improve accuracy; we therefore advocate request-level efficiency metrics that normalize throughput by tokens generated and trace an accuracy--speed frontier across reasoning efforts. We show that gpt-oss-puzzle-88B improves over gpt-oss-120B along the entire frontier, delivering up to 1.29$\times$ higher request-level efficiency.
Across various benchmarks, gpt-oss-puzzle-88B matches or slightly exceeds the parent on suite-average accuracy across reasoning efforts, with retention ranging from 100.8\% (high) to 108.2\% (low), showing that post-training architecture search can substantially reduce inference costs without sacrificing quality.
\end{abstract}

\section{Introduction}
\label{sec:intro}

Recent advances in large language models (LLMs) have been accompanied by a shift toward models that explicitly spend more computation at inference time \citep{gpt_o1, deepseek_r1, llama_nemotron}.
Reasoning-oriented models often operate over long contexts and generate long reasoning traces to improve answer quality, but the additional tokens dramatically increase serving cost.
In autoregressive inference, self-attention incurs computation that grows with sequence length, while the key-value (KV) cache introduces a memory footprint and bandwidth demand that scale linearly with the number of tokens.
At context lengths of 128K and beyond, KV-cache capacity and memory access can dominate end-to-end latency and throughput, limiting feasible batch sizes and lowering hardware utilization.
Agentic workflows can further amplify these costs by chaining many long-context calls and accumulating increasingly long histories.
These pressures motivate post-training inference optimization: tailoring a trained model to specific hardware and inference scenarios under explicit deployment constraints (e.g., memory footprint, latency, and throughput).

Reasoning introduces a second axis of efficiency---the number of tokens generated per request---so per-token throughput/latency can misstate end-to-end speedups (e.g., a 2$\times$ throughput gain is erased if traces grow 2$\times$).
Conversely, throughput gains can be reinvested into longer traces to improve accuracy.

\begin{figure}[t]
\centering
\begin{subfigure}[t]{0.499\textwidth}
  \centering
  \includegraphics[width=\linewidth]{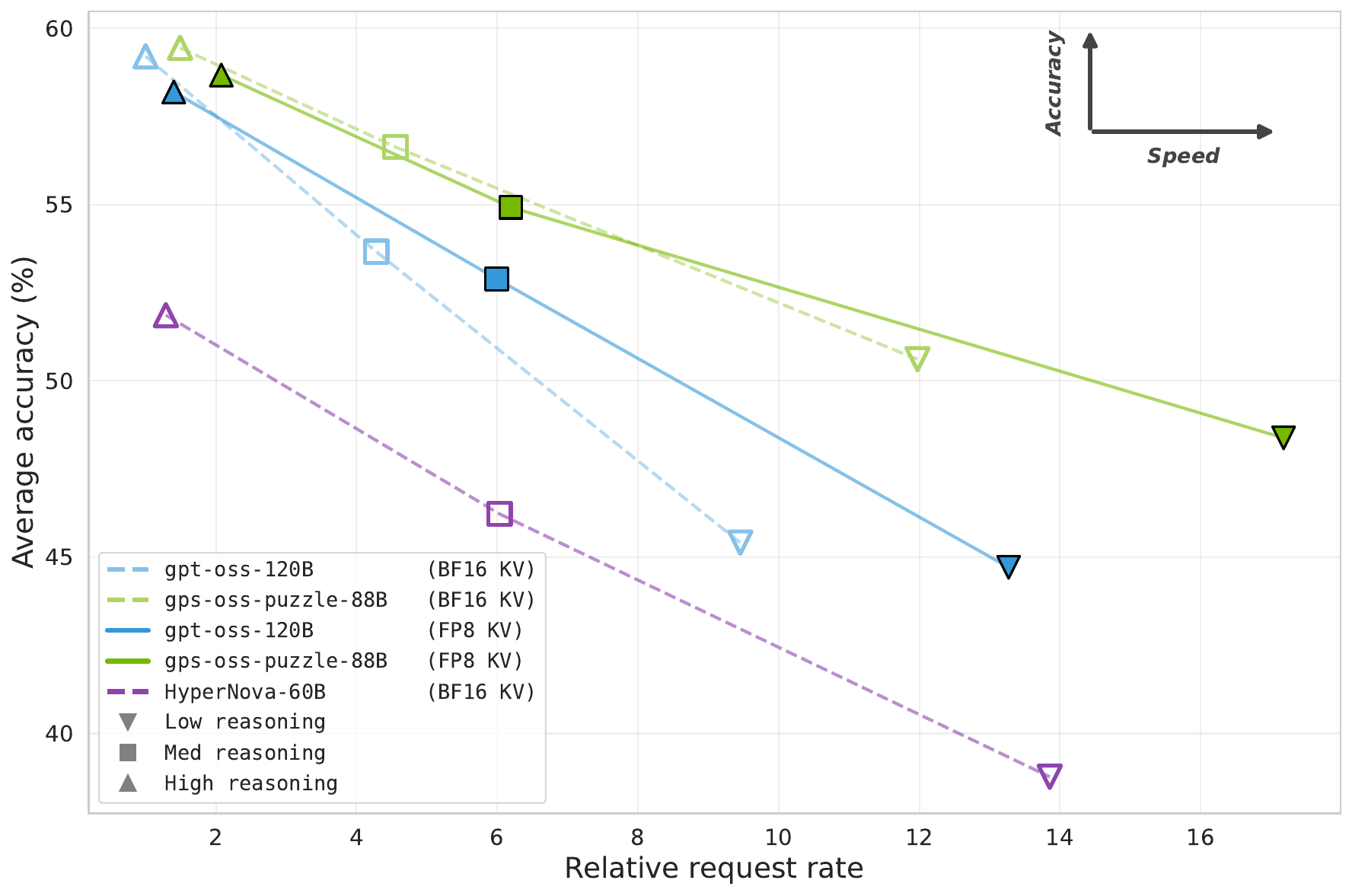}
  \caption{8$\times$H100 node}
\end{subfigure}\hfill
\begin{subfigure}[t]{0.499\textwidth}
  \centering
  \includegraphics[width=\linewidth]{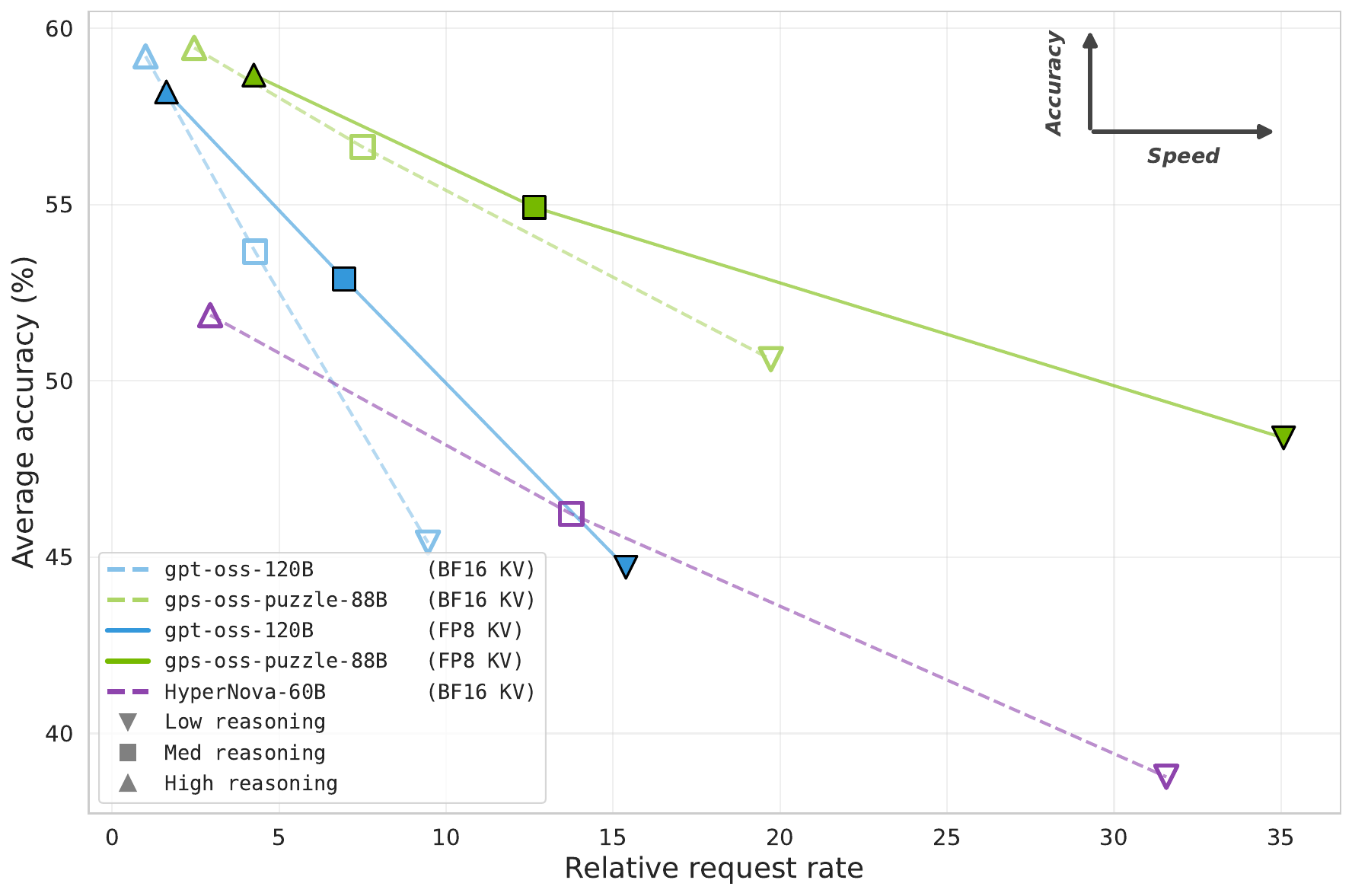}
  \caption{Single H100 GPU}
\end{subfigure}
\caption{Accuracy--speed frontier that accounts for both per-token throughput and tokens generated: (a) an 8$\times$H100 node and (b) a single H100 GPU. The x-axis shows \emph{relative request rate} (higher is faster), computed as max token throughput (best configuration per model) in a 64K/64K scenario divided by the average number of tokens generated per request across our benchmark suite, and normalized to gpt-oss-120B (KV BF16, high reasoning effort) in the corresponding hardware setting.
The y-axis is the suite's average accuracy.
Colors denote models (\textcolor{blue}{blue: gpt-oss-120B}; \textcolor{nvidia-green}{green: gpt-oss-puzzle-88B}; \textcolor{hypernova-purple}{purple: HyperNova-60B}~\citep{hypernova60b}), line style denotes KV precision (KV BF16 dashed; KV FP8 solid), and markers denote reasoning effort (High/Medium/Low).
HyperNova-60B is a third-party compressed derivative of gpt-oss-120B.}
\label{fig:accuracy_speed_frontier}
\end{figure}

Puzzle \citep{puzzle} is a decomposed neural architecture search (NAS) framework for improving LLM inference efficiency via neural architectural optimization.
Starting from a trained parent model, Puzzle constructs a \emph{block library}, a discrete set of per-layer alternatives (``puzzle pieces''), and associates each candidate block with measured resource costs in the target deployment scenario.
To make search tractable at LLM scale, Puzzle uses a decomposed \emph{replace-1-block} scoring scheme: each candidate block is evaluated in isolation by swapping it into the parent model at a single location, and the quality of a full architecture is estimated as a sum of its per-layer replace-1-block scores.
Given per-block costs and scores, Puzzle solves a mixed-integer program (MIP) to select one block per layer that maximizes estimated quality under deployment constraints, thus producing a heterogeneous architecture tailored to the target hardware and desired inference scenarios.
While the replacement process can be somewhat destructive (depending on compression levels) the reassembled model can be ``healed'' and refined with a short end-to-end knowledge distillation training phase to improve blocks' functionality and inter-block compatibility.
Puzzle enabled effective compression of Llama 3 Instruct models~\citep{llama3} in the Llama-Nemotron series, achieving $1.7\times$--$2.1\times$ speedup gains while maintaining competitive performance across benchmarks \citep{ffn-fusion,llama_nemotron}.

Applying Puzzle to recent LLMs that incorporate new architectural components and are expected to perform reasoning over longer generation sequences introduces new challenges.
First, mixture-of-experts (MoE) \citep{ShazeerMMDLHD17, FedusZS22} feed-forward networks (FFNs) are now widespread \citep{deepseekv3, qwen3, gpt_oss}, and practical MoE modifications must respect routing behavior and expert-parallel deployment constraints, making ``how much to prune'' a layer-dependent decision.
Second, long-context reasoning makes attention a dominant bottleneck and strongly incentivizes KV-cache-reducing mechanisms such as window/streaming attention \citep{longformer_window_attention, sink_attention_streaming_llm}; yet switching the wrong layers to window attention can harm long-range dependencies.
Finally, token-local replace-1-block scores (e.g., KL- or activation-MSE-based) may not reliably predict long-context degradations when switching to window attention, since they do not directly probe long-range interactions.

In this technical report, we cover the extension and application of Puzzle to the recent gpt-oss-120B model \citep{gpt_oss}, a strong open-weights reasoning model, and derive \textbf{gpt-oss-puzzle-88B}\footnote{Available on Hugging Face: \url{https://huggingface.co/nvidia/gpt-oss-puzzle-88B}.}, a deployment-optimized derivative optimized for both long- and short-context serving (Figure~\ref{fig:model_architecture}).
Our primary objective is deployment on an 8$\times$H100 node across both long- and short-context serving scenarios, with the long-context 128K setting being KV-cache--constrained.
The resulting model improves max throughput by 1.63$\times$ and 1.22$\times$ in long- and short-context settings, respectively, and after post-training matches or slightly improves the parent model's accuracy across the reasoning budgets. Our model also delivers throughput speedups of 2.82$\times$ on a single NVIDIA H100 GPU.
gpt-oss-puzzle-88B improves over gpt-oss-120B along the entire frontier, delivering up to 1.29$\times$ higher request-level efficiency with 8.2\%  improvement of relative average accuracy at low reasoning effort. The effort length ratio is maintained in the same range as the parent model, thereby preserving a reliable user facing control to trade cost for quality (Figure~\ref{fig:accuracy_speed_frontier}).
Since compression changes token counts across reasoning efforts, we also evaluate efficiency at the request level by accounting for tokens generated in practice, and track \emph{Effort Length Ratio}, defined as the ratio of generation lengths under high versus low effort. 

We run Puzzle once with a multi-constraint objective spanning both long- and short-context serving scenarios.
We score MoE pruning alternatives with an activation-based signal and score window-attention alternatives with a dedicated long-context reasoning signal. 
After selecting a compressed student model, we train it with knowledge distillation to recover quality lost from blockwise substitutions. We then perform reinforcement learning, training two complementary variants and merging them to further improve accuracy while keeping generation length low. Finally, we apply FP8 KV-cache quantization with max-calibrated KV scales to reduce KV footprint in the serving stack.

\paragraph{Our contributions are:}
\begin{itemize}
  \item Adapting Puzzle-style architecture search to MoE layers via heterogeneous expert removal under expert-parallel constraints.
  \item Identifying which attention layers can be converted to window attention using long-context-aware scoring, yielding large KV-cache savings while preserving capability.
  \item Using knowledge distillation to recover quality losses introduced during blockwise substitution.  
  \item Applying reinforcement learning to improve reasoning accuracy, training two complementary variants and merging them to maintain low generation length.
  \item Applying FP8 KV-cache quantization with calibrated KV scales, enabling \(\sim 2\times\) KV-cache token capacity and faster attention modules for long-context serving.
  \item Releasing \textbf{gpt-oss-puzzle-88B}, a deployment-optimized model derived from gpt-oss-120B.
\end{itemize}

The remainder of this paper is organized as follows: Section~\ref{sec:method} describes our Puzzle procedure; Section~\ref{sec:setup} details the training and evaluation setup; and Section~\ref{sec:results} presents throughput and accuracy results.

\section{Puzzle Optimization}
\label{sec:method}
Puzzle \citep{puzzle} is a decomposed \emph{neural architecture search} (NAS) framework for LLMs.
Given a trained ``\emph{parent model}'', Puzzle searches for a derivative architecture that satisfies deployment efficiency constraints (e.g., memory footprint, latency, and throughput) while preserving the parent’s accuracy.
It does so by (i) defining a discrete search space of alternative layer implementations pieces, (ii) estimating and assigning each alternative piece a quality score (comprising of its efficiency/accuracy profile), and (iii) solving a \emph{mixed-integer program} (MIP) to select one alternative piece per layer under the target constraints.

\textbf{Blocks, subblocks, and search space.}
Following \citep{puzzle}, we refer to a transformer layer as a \emph{block}, composed of two main \emph{subblocks}: the attention module and the feed-forward module (in gpt-oss-120B, an MoE FFN).
For each layer $i$, the search space combines an attention choice $\mathcal{A}_i = \{a_{i,1},\ldots,a_{i,m}\}$ with an FFN choice $\mathcal{F}_i = \{f_{i,1},\ldots,f_{i,n}\}$, yielding $\mathcal{A}_i \times \mathcal{F}_i$.
Because long-context inference is increasingly dominated by KV-cache size, our attention alternatives in this work focus on standard full-context attention and \emph{window attention} \citep{longformer_window_attention}. Window attention keeps the KV cache bounded by a fixed window size, making its memory cost insensitive to the total sequence length.
Our MoE FFN alternatives vary the number of experts, in a manner compatible with expert-parallel inference schemes.

\textbf{Replace-1-block scoring.}
To score the importance of each layer/subblock, we follow the activation-based scoring used in Nemotron-H \citep{nemotron-h}.
Given an input, we compute the intermediate activation tensor right before the LM head in the full parent model, and the corresponding tensor in a model where the particular layer is replaced with the alternative block variant. In our case, the alternative is an MoE subblock with a reduced number of experts.
Layer importance is then the mean squared error (MSE) between these two activation tensors.
We compute these MSE scores per sample, rank layers by importance, and average these rankings over a small random subset of the training data to obtain a reliable estimate of importance that takes into account sample variability.
These are called \emph{replace-1-block} scores \citep{puzzle}.
During search, candidate architectures are not evaluated directly; instead, their quality is estimated as the sum of the replace-1-block scores of their chosen subblocks.

\textbf{Optimized Scenarios.}
We run Puzzle once with a multi-constraint objective that targets two deployment scenarios on a single 8$\times$H100 node: a long-context 64K/64K scenario that requires a 1.6$\times$ throughput improvement over the gpt-oss-120B parent, and a short-context 4K/4K scenario that requires a 1.2$\times$ improvement.
Meeting these constraints pushes the optimization toward different architectural levers: for 64K/64K, where KV-cache I/O dominates decoding, Puzzle primarily improves efficiency by converting 8 out of the 18 global attention layers into window attention layers, achieving a KV-cache size that is 40\% smaller compared to the parent model gpt-oss-120B. 
For 4K/4K, where KV-cache pressure is significantly lower, Puzzle primarily relies on MoE expert pruning (removing 25\% of experts) to meet the target.
Although the constraints are specified at the 8-GPU node level, they also translate into large gains on a single H100 GPU: as shown in Figure~\ref{fig:ablation_scaling}, the parent model is often memory-limited to batch size 1, while expert pruning and window attention free memory that enables larger effective batch sizes, leading gpt-oss-puzzle-88B to achieve a $2.82\times$ speedup on a single H100 GPU in a 64K/64K scenario, and a $2.44\times$ speedup in a 4K/4K scenario.
In the following paragraphs, we detail the two main architectural components, MoE expert pruning and selective window attention, and how we implement and score them.

\begin{figure}[b]
\centering
\includegraphics[width=1.0\textwidth]{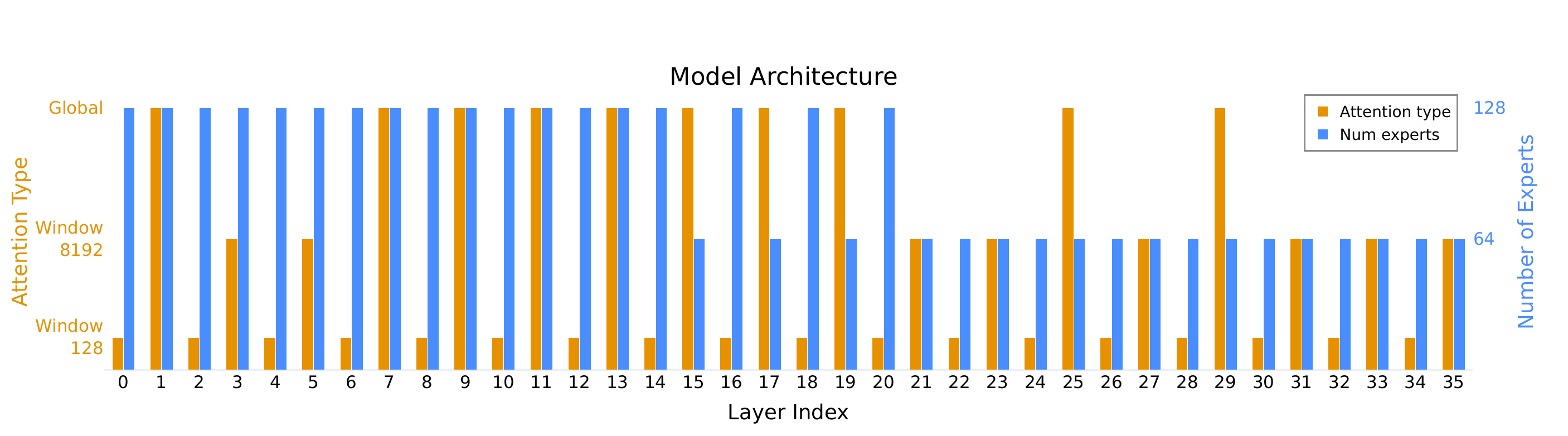}
\caption{Our model architecture as chosen by Puzzle. Note how the earlier MoE layers appear to be far more important than the later ones. The parent model gpt-oss-120B has 128 experts per layer and alternates between sliding window attention with 128 tokens and global attention.}
\label{fig:model_architecture}
\end{figure}

\paragraph{Pruning mixture-of-experts layers.}
Our main method to tackle MoE layers is \emph{heterogeneous expert removal}. For each MoE layer, we first rank the experts by their contribution to the output of the MoE module over a validation set, then construct a block library made of multiple variants of this layer, each keeping a different number of the original experts from the parent layer: 8, 16, 32, 64, 96 or 128 of the original 128 experts. Using this block library, we utilize Puzzle to determine how drastically each MoE layer in the model should be pruned, resulting in a heterogeneous architecture where the size of each MoE layer is determined by its estimated impact on model quality. Note how the earlier MoE layers appear to be far more important than the later ones (Figure~\ref{fig:model_architecture}).

To rank the experts within each layer, we calculate expert contribution scores. We propagate samples through the model, and for each MoE layer we compare the output of the original layer vs the output of this layer with one expert removed: $\text{expert\_score}_i = \mathbb{E}_{x}\!\left[\mathrm{MSE}\bigl(f(x), f^{(i)}(x)\bigr)\right]$, where $f$ is an MoE layer, $f^{(i)}$ is $f$ with expert $i$ removed, and $x$ are the input hidden states to layer $f$.
We found that using data compatible with the parent's distribution is very important for correct importance estimation, and therefore used our generated LNPT-gpt-oss dataset (Section ~\ref{paragraph:dataset}). To calculate expert contribution scores we used 12K packed sequences of length 8K tokens each, and to calculate replace-1-block scores we used 128 packed sequences of length 32K tokens each.

\paragraph{Window attention for long-context inference.}
In long-context settings, the attention subblock becomes a primary bottleneck due to its dependence on the sequence length $L$: both the computation of attention scores and the storage/access of the key--value (KV) cache scale unfavorably as $L$ grows.
To mitigate this, several recent models adopt a mix of global attention layers and sliding window attention layers, including Gemma~\citep{gemma2,gemma3} and gpt-oss \citep{gpt_oss}.

Window attention restricts each token to attend only to the most recent $W$ tokens, rather than the entire prefix. This modification reduces the effective attention span from $L$ to $W$, yielding substantial savings in both compute and KV-cache memory IO. In particular, the per-layer KV-cache footprint becomes proportional to $W$ (instead of $L$), and the attention score computation similarly depends on $W$, which is especially beneficial when serving very long sequences.

Previous works using Puzzle \citep{puzzle, ffn_fusion, llama_nemotron} reported strong long-context retention even without dedicated long-context uptraining \citep{puzzle}. However, those efforts mainly considered attention alternatives that preserve the global receptive field, such as \emph{grouped-query attention} (GQA), or, in extreme cases, removing an attention layer altogether (which probably indicates that this attention layer contributed very little to begin with).
While GQA reduces the KV-cache size by sharing key/value heads across query heads, its KV cache still grows linearly with sequence length.
Window attention, in contrast, makes the KV cache effectively fixed-size and is therefore particularly attractive at very long contexts.

However, converting \emph{all} full-context attention layers to window attention typically incurs a serious quality degradation, since some layers rely on global context to support long-range dependencies. Therefore, our goal is to identify a \emph{subset} of attention layers that are most amenable to window attention (i.e., ``window-able'') and apply window attention only to those layers, while keeping the remaining layers as full-context attention. This selective replacement enables large efficiency gains in long-context scenarios while preserving the accuracy of the parent model.

Our parent model, gpt-oss-120B, alternates between sliding window attention with 128 tokens and global attention. Our attention block library included all the original parent attention blocks, and an additional alternative block for each original global attention layer, where it is replaced by a sliding window attention layer with 8192 tokens.
To determine which global attention layers are the most suitable to be transformed into window attention, we needed a dedicated scoring method that is long-context aware. While it is robust to a large variety of layer types and pruning methods, our standard method for calculating replace-1-block scores has an inherent locality bias, as it compares the model's final hidden states used for next-token prediction, and language modeling is an inherently local task for most tokens.
To capture long-term dependencies, when constructing the block library for attention layers, we instead measured accuracy scores on the Artificial Analysis Long-Context Reasoning benchmark (AA-LCR~\citep{artificialanalysis2025methodology}) for each block replacement. We conduct an ablation experiment comparing the two types of scoring in Appendix~\ref{ablation_scoring}.

To further support long-context behavior after these architectural changes, we modify the YaRN positional encoding by increasing the RoPE scaling factor from 32 to 56. Although 32 matches the nominal 4K $\rightarrow$ 128K extension ratio used by the parent model, YaRN applies frequency-dependent scaling; at 128K, parts of the transition band can still experience substantial phase wrapping. Using factor 56 increases the effective RoPE periods and yields more stable long-range attention, improving long-context retrieval at 128K. We note that the long-context accuracy of the parent model may also be improved by tuning the RoPE scaling factor.

\section{Training, Quantization and Evaluation Setup}
\label{sec:setup}

\paragraph{Dataset.}
\label{paragraph:dataset}
For expert contribution scores, MoE replace-1-block scores, knowledge distillation, and quantization, we used prompts from the \href{https://huggingface.co/datasets/nvidia/Llama-Nemotron-Post-Training-Dataset}{nvidia/Llama-Nemotron-Post-Training-Dataset}~\citep{llama_nemotron}, which is designed to enhance performance in mathematics, code generation, general reasoning, and instruction following. For each prompt, we generated responses from the parent model under both \texttt{high} and \texttt{medium} reasoning effort settings. We refer to the result as the \emph{LNPT-gpt-oss} dataset. Block scoring and quantization were performed exclusively on the \texttt{high} reasoning effort subset, whereas knowledge distillation employed an equal mixture of \texttt{high} and \texttt{medium} effort responses to preserve accuracy across diverse usage patterns.

\paragraph{Knowledge Distillation.}
Following the puzzle phase, we trained the model using a knowledge distillation objective on the LNPT-gpt-oss dataset for a total of 84B tokens, with the goal of improving inter-block compatibility and recovering any quality degradation introduced by blockwise substitution. During this stage, the MoE experts and router were kept frozen. Training was performed with a sequence length of 128K and a global batch size of 33M tokens, using the Megatron-LM framework~\citep{megatron-lm}.

\paragraph{Reinforcement Learning.}
Following distillation, we run a reinforcement learning stage to further improve reasoning accuracy, while tracking how training choices affect generation length, since tokens per request directly determine end to end serving cost. We build on the repositories, multi-environment RL framework, and datasets of \citet{nemotron3}, training on the mathematics, coding, and general reasoning environments while excluding tool-use environments from the mixture. During this stage, the MoE experts and router were kept frozen. Across runs, we keep the recipe fixed (including a constant learning rate of $1\text{e-}6$) and vary only the reasoning-effort composition of the RL data.

Varying this single axis reveals the central tension of the RL stage: Training exclusively on high-effort reasoning data reliably improves reasoning accuracy, yet it lengthens generations across all efforts and shifts effort length ratio away from the teacher range, making high effort disproportionately expensive (Table~\ref{tab:token_count}). At the other extreme, training on a balanced mixture of efforts acts as an implicit length regularizer, leading to lower verbosity at high reasoning effort. Generation length across efforts regress toward a shared mean and effort becomes less effective at steering generation length, weakening controllability. This regime also underperforms in accuracy at the same budget, suggesting it requires substantially more iterations to match the high effort focused setting.
To reconcile these behaviors, we combine the two RL candidates via checkpoint weight averaging. The resulting model preserves near peak reasoning accuracy from the high effort trained policy while substantially reducing verbosity and restoring effort length ratio toward the teacher range, so users can reliably trade cost for quality.

\paragraph{KV Quantization.}
\label{sec:quantization}

In addition to keeping the MXFP4 quantization of the experts as in the parent model gpt-oss-120B, we employ FP8 KV quantization to reduce the memory footprint of the KV cache. This enabled both $2\times$ token capacity in the KV cache and faster attention modules, which is especially beneficial in long context scenarios, such as our case with reasoning models.
We first tried the common practice of not using KV scales. However, this resulted in subpar accuracy.
Therefore, we computed KV scales based on our LNPT-gpt-oss dataset using max calibration, resulting in better accuracy. 
Furthermore, we rounded the scales up to a power of 2. This makes sure that rounding errors behave similarly to not having scales, but with a better adaptation of the dynamic range to mitigate underflow errors (all scales are below 1, so overflow errors seem to be less of a concern).
Our inference benchmarks can be found in Section \ref{sec:bench} and our accuracy benchmarks can be found in Section \ref{sec:accuracy}.

\paragraph{Accuracy evaluations.}
We focused on reasoning benchmarks for accuracy evaluations. 
We used MMLU-Pro \citep{mmlu-pro} and HLE \citep{hle} for reasoning and knowledge, GPQA-Diamond \citep{gpqa} for scientific reasoning, AIME-25 \citep{maa_aime} for mathematical knowledge,
SciCode \citep{scicode} for coding and IFBench \citep{ifbench} for general instruction following. Since we have used AA-LCR scores internally as a part of the Puzzle process, we're using RULER \citep{ruler} results as another benchmark to measure long context performance. We followed OpenAI's reasoning efforts scheme and evaluated our model using \texttt{high}, \texttt{medium} and \texttt{low} reasoning efforts. In order to reduce variance in accuracy we calculated each benchmark several times and report averaged results. 

We have used the Nemo-Skills repo \cite{nemo_skills2024} to run these benchmarks. Evaluation parameters are listed in Appendix~\ref{sec:eval_parameters}. 

\paragraph{Inference efficiency benchmarking.}
For each serving scenario we sweep tensor-parallel degree (TP $\in \{1,2,4,8\}$) and a grid of batch sizes.
For max-throughput numbers we select the best configuration \emph{per model} (instead of fixing TP), ensuring a fair comparison. Each configuration is measured 3 times.
The plots include $\pm 1\sigma$ bands to visualize uncertainty.

\section{Results}
\label{sec:results}
\begin{figure}[bt]
\centering
\includegraphics[width=1.0\textwidth]{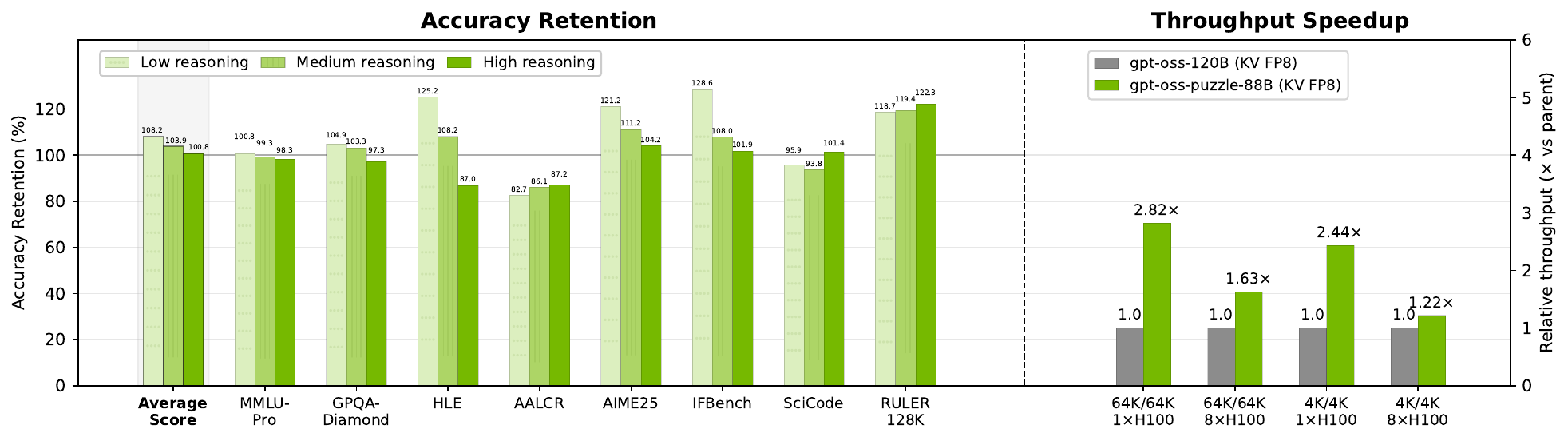}
\caption{Accuracy and throughput comparisons of gpt-oss-puzzle-88B with its parent, gpt-oss-120B (both with KV FP8).}
\label{fig:results_overview}
\end{figure}

\subsection{Inference Efficiency}
\label{sec:bench}
\paragraph{Why throughput (tok/s) is not enough for reasoning models.}
Per-token throughput (tok/s) and latency (ms/token) capture architectural efficiency, but for reasoning models they are not sufficient: different model variants and reasoning-effort modes can generate different numbers of tokens per request, directly changing end-to-end latency and cost.
Accordingly, we distinguish between (i) \emph{token throughput} under fixed input/output lengths (this subsection) and (ii) \emph{request-level efficiency} that also accounts for tokens generated in practice.
Figure~\ref{fig:accuracy_speed_frontier} summarizes this request-level trade-off by plotting accuracy against a \emph{relative request rate}, computed as the max token throughput (best serving configuration per model) divided by the average tokens generated per request (and normalized to the gpt-oss-120B KV BF16 high-effort baseline); Appendix~\ref{app:frontier_raw_results} reports the raw numbers underlying the figure.
As an illustrative example, HyperNova-60B~\citep{hypernova60b}\footnote{We evaluated HyperNova-60B only with KV FP16, using the Hugging Face checkpoint available as of Feb.\ 11, 2026.} is a third-party compressed derivative of gpt-oss-120B that targets architectural efficiency, but it generates longer reasoning traces, which can erase per-token throughput gains at the request level.
Table~\ref{tab:token_count} reports complementary generated-token statistics for our models.
With this framing in mind, the remainder of this subsection focuses on architectural speedups: maximum token throughput and throughput--latency trade-offs in fixed 4K/4K and 64K/64K serving scenarios.
We compare our optimized \textbf{gpt-oss-puzzle-88B} against the \textbf{gpt-oss-120B} parent.
We focus on the specific constraints used during the optimization process and highlight the quantized gpt-oss-puzzle-88B configuration as the final optimized serving setup.

\paragraph{Optimization Constraints.}
\begin{itemize}
    \item \textbf{Short Context (4K/4K):} A target improvement of \textbf{20\%}. This scenario is typically MoE-dominant, driving the optimization of the Mixture-of-Experts (MoE) layers (expert pruning).
    \item \textbf{Long Context (64K/64K):} A target improvement of \textbf{60\%}. This scenario is KV-cache-dominant, driving the optimization of the attention mechanism (selective window attention).
\end{itemize}

\paragraph{Hardware.} We report results on NVIDIA H100 (8$\times$ H100 80GB HBM3 node).
We first present a detailed analysis on H100, with a B200 case study in Appendix~\ref{sec:b200_case_study}.

\paragraph{Scenarios and notation.}
We denote inference scenarios as \emph{input/output} token lengths (e.g., \textbf{64K/64K} means a 64K-token prompt followed by generating 64K tokens).
The first number primarily stresses the \emph{prefill} phase, while the second stresses the \emph{decode} phase.

\subsubsection{H100 Results: Meeting the Dual Constraints}

Table~\ref{tab:inference_headline_scenarios} summarizes the speedups across our target deployment scenarios on a single node.
The results demonstrate that \textbf{gpt-oss-puzzle-88B} exceeds both optimization targets:
\begin{enumerate}
    \item In the \textbf{4K/4K} scenario, we achieve a \textbf{1.22$\times$} speedup, validating the efficiency gains from expert pruning in MoE dominant regimes.
    \item In the \textbf{64K/64K} scenario, we achieve a \textbf{1.63$\times$} speedup, validating the impact of selective window attention in kv cache dominant regimes.
\end{enumerate}

\begin{table}[htbp]
\centering
\small
\caption{Inference scenarios (H100): gpt-oss-puzzle-88B vs gpt-oss-120B parent.}
\label{tab:inference_headline_scenarios}
\resizebox{1.0\linewidth}{!}{%
\begin{tabular}{llp{4.2cm}p{4.2cm}r}
\toprule
Scenario & Description & \textbf{gpt-oss-puzzle-88B} & gpt-oss-120B & Speedup \\
\midrule

4K/4K & Max throughput on 8$\times$H100 node & \textbf{36.1K tok/s} & 29.6K tok/s & 1.22$\times$ \\
64K/64K & Max throughput on 8$\times$H100 node & \textbf{9.3K tok/s} & 5.7K tok/s & 1.63$\times$ \\

\midrule

4K/4K & Max throughput on single H100 & \textbf{3.3K tok/s} & 1.4K tok/s & 2.44$\times$ \\
64K/64K & Max throughput on single H100 & \textbf{0.8K tok/s} & 0.3K tok/s & 2.82$\times$ \\
\bottomrule

\end{tabular}
}
\end{table}

\begin{figure}[htbp]
    \centering
    \begin{subfigure}[b]{0.48\textwidth}
        \centering
            \includegraphics[width=\linewidth]{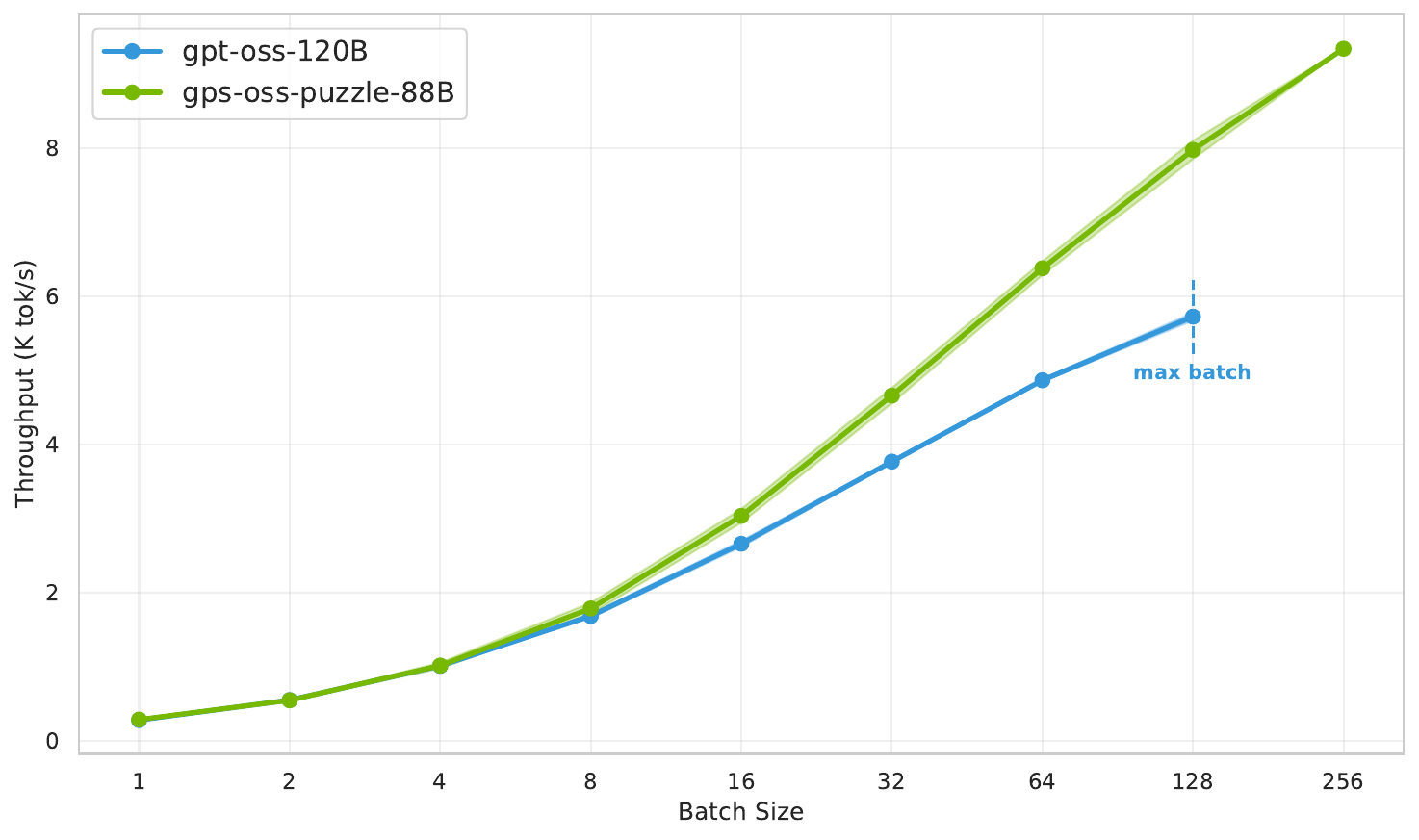}
        \caption{Long Context (64K/64K)}
        \label{fig:throughput_64K}
    \end{subfigure}
    \hfill
    \begin{subfigure}[b]{0.48\textwidth}
        \centering
        \includegraphics[width=\linewidth]{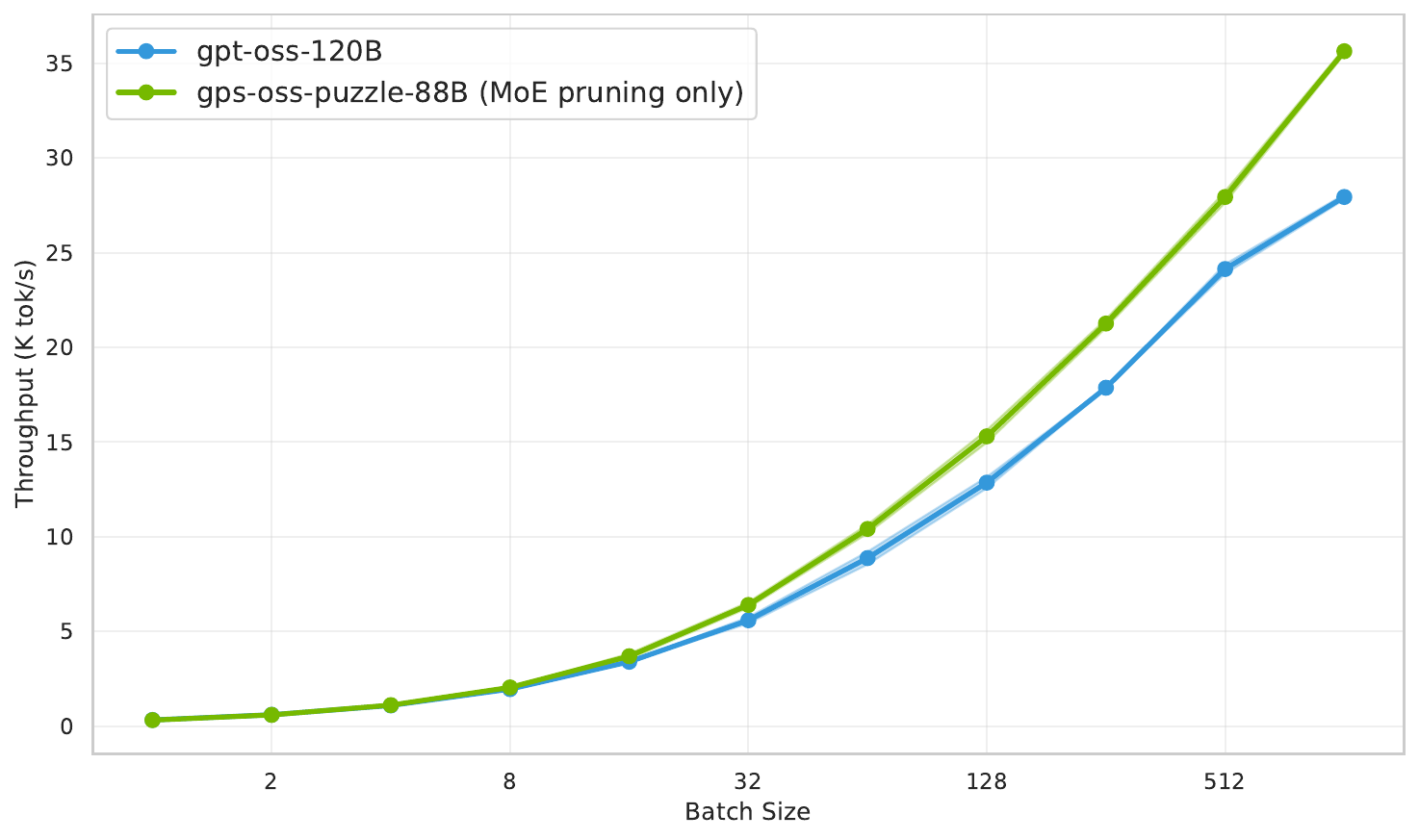}
        \caption{Short Context (4K/4K)}
        \label{fig:throughput_4K}
    \end{subfigure}
    \caption{Throughput scaling with batch size on an $8\times$ H100 node. Comparison of (a) long-context and (b) short-context scenarios. Both models (gpt-oss-120B and gpt-oss-puzzle-88B) use KV FP8. Shaded bands show $\pm 1\sigma$.}
    \label{fig:throughput_scaling}
\end{figure}

\paragraph{Scaling and trade-offs.}
Figure~\ref{fig:throughput_scaling} shows throughput scaling with batch size.
Because gpt-oss-puzzle-88B reduces both weight bandwidth and KV-cache footprint, it sustains scaling to higher batch sizes, improving GPU utilization and reaching peak throughput across both MoE-dominant (4K/4K) and KV-dominant (64K/64K) regimes.

\begin{figure}[htbp]
\centering
\includegraphics[width=0.92\textwidth]{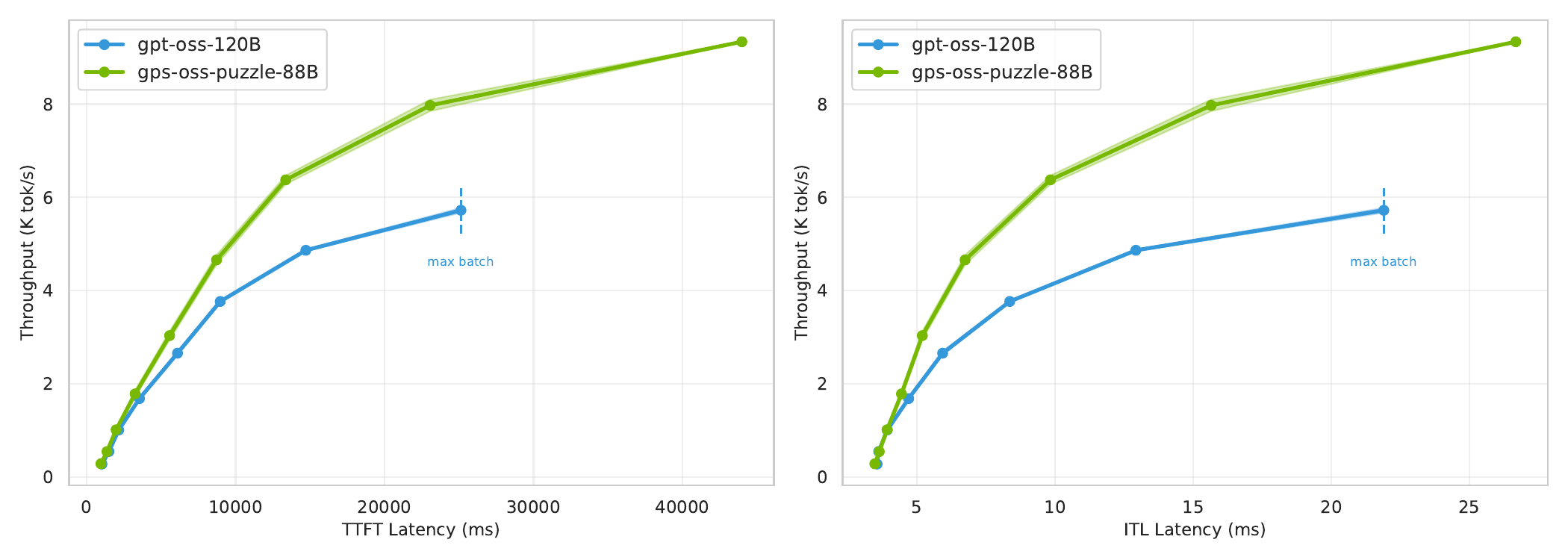}
\caption{Latency vs throughput trade-off (64K/64K). Both models (gpt-oss-120B and gpt-oss-puzzle-88B) use KV FP8. Shaded bands show $\pm 1\sigma$ across repeated runs.}
\label{fig:latency_64K}
\end{figure}

Figure~\ref{fig:latency_64K} illustrates the throughput--latency frontier for the 64K/64K scenario.
The optimized model offers superior trade-offs; for example, when constrained to a decode latency (ITL) of 10ms, gpt-oss-puzzle-88B delivers approximately \textbf{1.5$\times$} higher throughput (6.4K vs 4.2K tok/s) compared to the parent model.

\begin{figure}[htbp]
    \centering
    \includegraphics[width=0.7\textwidth]{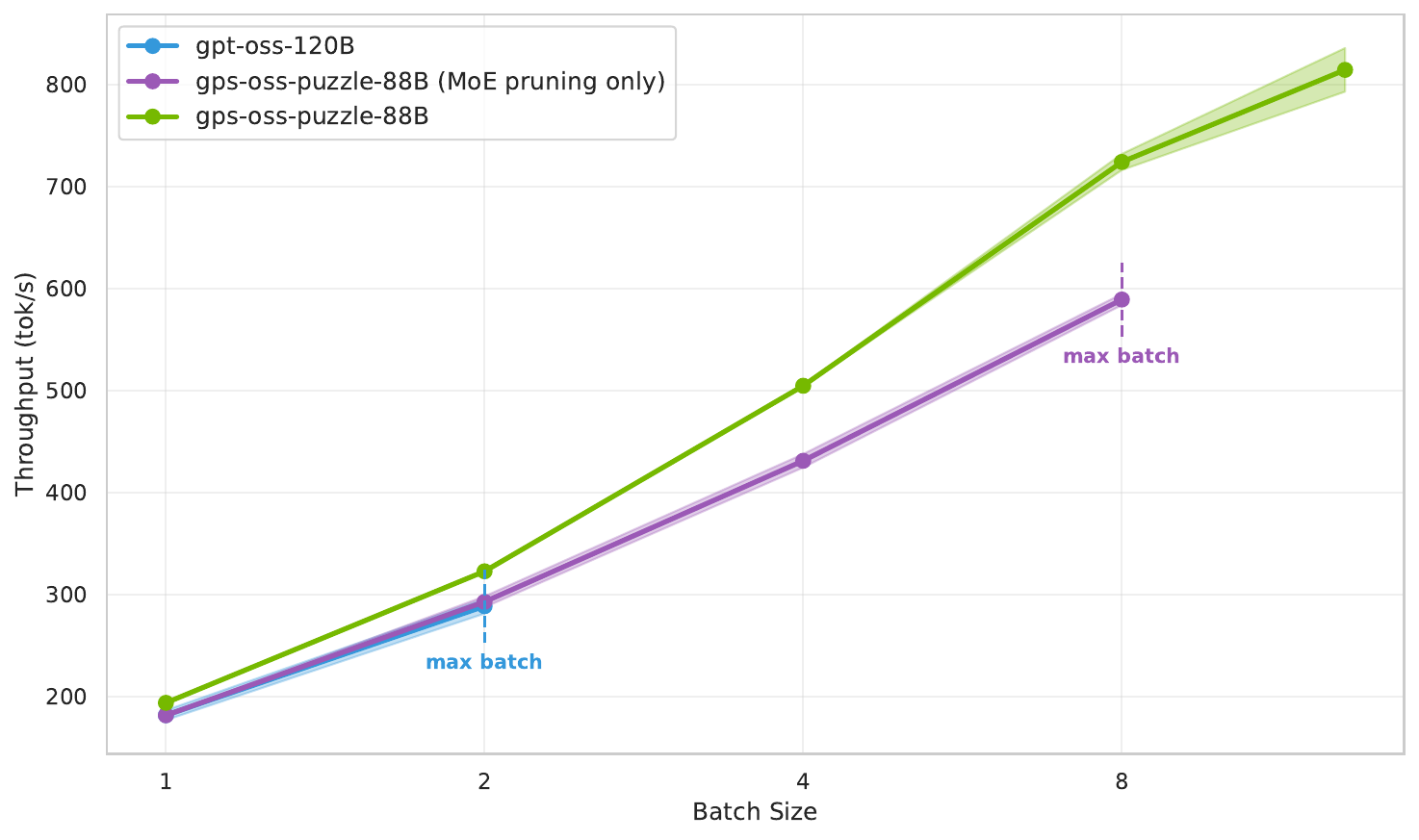}
    \caption{Single-H100 (TP=1) throughput scaling with batch size in the 64K/64K scenario. Compared to the gpt-oss-120B parent, MoE expert pruning and selective window attention reduce weight/KV-cache footprint and enable larger maximum batch sizes, improving utilization and throughput. All runs use KV FP8; shaded bands show $\pm 1\sigma$, and dashed markers indicate the maximum batch size that fits in memory.}
    \label{fig:ablation_scaling}
\end{figure}

\paragraph{Single GPU Efficiency.}
While our primary optimization constraints focused on full-node deployment, the architectural efficiency of gpt-oss-puzzle-88B is even more pronounced on constrained resources.
On a single H100 GPU, we observe improvements of \textbf{2.44$\times$} in the 4K/4K scenario and \textbf{2.82$\times$} in the 64K/64K scenario compared to the parent model.
Figure~\ref{fig:ablation_scaling} shows that the parent model is quickly memory-limited as batch size increases, while expert pruning and selective window attention free memory that enables larger effective batch sizes and higher throughput.
These single-device gains highlight the model's ability to operate effectively under strict memory capacity limits where the parent model struggles with batch scaling.

\FloatBarrier

\subsection{Accuracy Benchmarks}
\label{sec:accuracy}

We report accuracy results for our quantized and non-quantized gpt-oss-puzzle-88B models compared to the parent (gpt-oss-120B) in all reasoning efforts reported by OpenAI (low, medium and high) in Table~\ref{tab:accuracy}. 

From Table~\ref{tab:accuracy}, comparing the KV FP8 variants (our final serving configuration), gpt-oss-puzzle-88B achieves \(100.8\%\), \(103.9\%\), and \(108.2\%\) of the parent model's suite-average accuracy at high, medium, and low reasoning effort, respectively, despite having $\sim$73\% of the parent's total parameter count. Retention varies across benchmarks, and some tasks remain below the parent.

\begin{table}[htbp]
\centering
\small
\caption{Accuracy (\%) for gpt-oss-120B and gpt-oss-puzzle-88B (KV BF16/KV FP8) at low, medium, and high reasoning effort.}
\label{tab:accuracy}
\setlength{\tabcolsep}{3pt}
\begin{tabular}{lccccccccc}
\toprule
Model & \textbf{\makecell{Average\\Accuracy}} & \makecell{MMLU- \\Pro} & \makecell{GPQA- \\Diamond} & \makecell{HLE} & \makecell{AALCR} & \makecell{AIME25} & \makecell{IFBench} & \makecell{SciCode} & \makecell{RULER \\128K} \\
\midrule
\textbf{High reasoning} \\
gpt-oss-120B (KV BF16) & 59.20 & 80.41 & 77.78 & 18.16 & 48.75 & 91.46 & 64.46 & 41.72 & 50.89 \\
gpt-oss-120B (KV FP8) & 58.19 & 80.60 & 77.34 & 18.86 & 46.75 & 89.58 & 65.76 & 40.83 & 45.82 \\
gpt-oss-puzzle-88B (KV BF16) & 59.44 & 79.32 & 75.13 & 17.52 & 42.25 & 92.92 & 67.77 & 40.83 & 59.80 \\
gpt-oss-puzzle-88B (KV FP8) & 58.67 & 79.19 & 75.25 & 16.40 & 40.75 & 93.33 & 67.01 & 41.42 & 56.02 \\
\midrule
\textbf{Medium reasoning} \\
gpt-oss-120B (KV BF16) & 53.66 & 78.86 & 71.28 & 10.06 & 39.25 & 76.88 & 56.55 & 41.42 & 55.01 \\
gpt-oss-120B (KV FP8) & 52.89 & 78.71 & 68.88 & 9.68 & 41.50 & 77.92 & 58.67 & 42.75 & 45.04 \\
gpt-oss-puzzle-88B (KV BF16) & 56.64 & 78.03 & 69.70 & 10.57 & 36.00 & 86.88 & 65.56 & 39.64 & 66.71 \\
gpt-oss-puzzle-88B (KV FP8) & 54.93 & 78.18 & 71.15 & 10.47 & 35.75 & 86.67 & 63.35 & 40.09 & 53.77 \\
\midrule
\textbf{Low reasoning} \\
gpt-oss-120B (KV BF16) & 45.41 & 75.18 & 62.75 & 4.17 & 35.25 & 50.00 & 44.13 & 39.20 & 52.61 \\
gpt-oss-120B (KV FP8) & 44.71 & 75.04 & 60.54 & 4.40 & 34.75 & 51.88 & 43.71 & 40.09 & 47.28 \\
gpt-oss-puzzle-88B (KV BF16) & 50.61 & 75.56 & 64.77 & 5.33 & 31.75 & 66.25 & 56.38 & 38.17 & 66.70 \\
gpt-oss-puzzle-88B (KV FP8) & 48.38 & 75.62 & 63.51 & 5.51 & 28.75 & 62.89 & 56.21 & 38.46 & 56.11 \\
\bottomrule
\end{tabular}
\end{table}

\FloatBarrier

\paragraph{Generated Token Count Analysis} 
Table~\ref{tab:token_count} reports the average number of generated tokens across seven benchmarks for multiple reasoning efforts, comparing RL intermediate checkpoints, the final gpt-oss-puzzle-88B, and its parent gpt-oss-120B. Notably, the final gpt-oss-puzzle-88B produces slightly fewer tokens than the average for the RL variants, and its effort length ratio remains close to the parent. Despite higher token counts than the parent, gpt-oss-puzzle-88B is more efficient at the request level due to architectural throughput gains, as seen in Figure~\ref{fig:accuracy_speed_frontier}.
\begin{table}[t]
\centering\small
\begin{center}
\small
\setlength{\tabcolsep}{3pt}
\setlength{\extrarowheight}{2pt}
\begin{tabular}{lccc|ccc}
\hline
& \multicolumn{3}{c|}{\textbf{RL Variants}} & & & \\
\cline{2-4}
\textbf{Reasoning Effort} & \textbf{Pre-RL} & \textbf{Balanced Mix} & \textbf{High} & \textbf{gpt-oss-puzzle-88B} & \textbf{gpt-oss-120B} & \textbf{Ratio $\sfrac{\mathrm{Ours}}{\mathrm{Parent}}$} \\
\hline
High   & 15.63 & 9.96 & 21.56 & 14.28 & 13.05 & $1.09\times$ \\
Medium & 3.2 & 3.88 & 6.30  & 4.79  & 3.05  & $1.57\times$ \\
Low    & 1.4 & 2.01 & 1.56  & 1.73  & 1.35  & $1.28\times$ \\
\hline
\end{tabular}
\end{center}
\caption{Generated tokens (K) by reasoning effort, across RL checkpoints (Pre-RL, Balanced Mix - RL on all efforts, High - RL on high-effort only) and final models (gpt-oss-puzzle-88B vs parent gpt-oss-120B, with ratio). Averaged over MMLU-Pro, HLE, GPQA-Diamond, AIME-25, IFBench, SciCode, and AALCR. All models use KV FP8.}
\label{tab:token_count}
\end{table}

\FloatBarrier

\subsubsection{KV Scaling Impact}
As discussed in Section \ref{sec:quantization}, we found that using KV scales gives better results. In Table \ref{tab:accuracy_quant}, we detail the accuracy results for our KV quantization experiments.

\begin{table}[htbp]
\centering
\small
\caption{Accuracy results for the gpt-oss-puzzle-88B model when using FP8 KV quantization with and without KV scales}

\label{tab:accuracy_quant}





















\setlength{\tabcolsep}{3pt}
\begin{tabular}{lccccccccc}
\toprule
Model & \textbf{\makecell{Average\\Accuracy}} & \makecell{MMLU- \\Pro} & \makecell{GPQA- \\Diamond} & \makecell{HLE} & \makecell{AALCR} & \makecell{AIME25} & \makecell{IFBench} & \makecell{SciCode} & \makecell{RULER \\128K}  \\

\midrule
\textbf{High reasoning} \\
No Scales  & 58.14 & 79.40 & 75.00 & 15.99 & 40.25 & 90.42 & 66.89 & 40.68 & 56.47 \\
KV Scales  & 58.67 & 79.19 & 75.25 & 16.40 & 40.75 & 93.33 & 67.01 & 41.42 & 56.02 \\
\midrule
\textbf{Medium reasoning} \\
No Scales  & 55.47 & 77.71 & 70.58 & 11.54 & 35.50 & 87.08 & 64.97 & 41.12 & 55.25 \\
KV Scales  & 54.93 & 78.18 & 71.15 & 10.47 & 35.75 & 86.67 & 63.35 & 40.09 & 53.77 \\
\midrule
\textbf{Low reasoning} \\
No Scales  & 48.01 & 75.46 & 63.19 & 5.14 & 26.25 & 65.21 & 55.02 & 38.61 & 55.17 \\
KV Scales  & 48.38 & 75.62 & 63.51 & 5.51 & 28.75 & 62.89 & 56.21 & 38.46 & 56.11 \\
\bottomrule
\end{tabular}    
\end{table}

\FloatBarrier

\section{Discussion}
\label{sec:discussion}

This technical report introduced an extension to Puzzle, and applied it to optimize the gpt-oss-120B parent model. Starting from gpt-oss-120B, we derived gpt-oss-puzzle-88B by jointly targeting an 8$\times$H100-node throughput improvement of 1.6$\times$ in the long-context 64K/64K scenario and 1.2$\times$ in the short-context 4K/4K scenario. gpt-oss-puzzle-88B achieves significant throughput improvement over the parent while matching the parent on suite-average accuracy across reasoning efforts (Figure~\ref{fig:results_overview}). At the request level, which also accounts for tokens generated across reasoning efforts, gpt-oss-puzzle-88B improves along the accuracy--speed frontier (Figure~\ref{fig:accuracy_speed_frontier}).

Our approach combines several complementary ingredients: (i) \emph{heterogeneous MoE expert pruning} in which layers are pruned to different expert counts based on activation-based replace-1-block scores, (ii) \emph{selective window attention} replacements scored with a long-context benchmark signal to preserve long-range behaviors, (iii) training complementary variants using reinforcement learning and merging them to further improve accuracy while keeping generation length low and (iv) FP8 KV-cache quantization with calibrated scales to further reduce KV-cache footprint. Together, these results show that post-training architecture search can substantially reduce the cost of both long- and short-context serving while matching or even improving quality.

\bibliography{ref}
\bibliographystyle{plainnat}

\newpage
\appendix
\onecolumn

\section{B200 Inference Efficiency Case Study: 64K/64K}
\label{sec:b200_case_study}

Table~\ref{tab:inference_b200} summarizes full-node inference throughput on an 8$\times$B200 system across tested scenarios.
In the 64K/64K setting, gpt-oss-puzzle achieves up to \textbf{1.4$\times$} higher throughput than the parent gpt-oss-120B.

\begin{table}[htbp]
\centering
\small
\caption{B200 full-node (8$\times$ B200) inference throughput: gpt-oss-puzzle-88B vs Parent. Values are max throughput at the best configuration (TP/BS omitted for readability).}
\label{tab:inference_b200}
\begin{tabular}{lrrrr}
\toprule
Scenario & \textbf{gpt-oss-puzzle-88B} & gpt-oss-120B (parent) & gpt-oss-puzzle-88B/Parent \\
\midrule
64K/64K &  \textbf{21.2K tok/s} & 15.2K tok/s & 1.40$\times$ \\
4K/4K & \textbf{73.6K tok/s} & 63.9K tok/s & 1.15$\times$ \\
\bottomrule
\end{tabular}

\end{table}

\begin{figure}[htbp]
\centering
\includegraphics[width=0.92\textwidth]{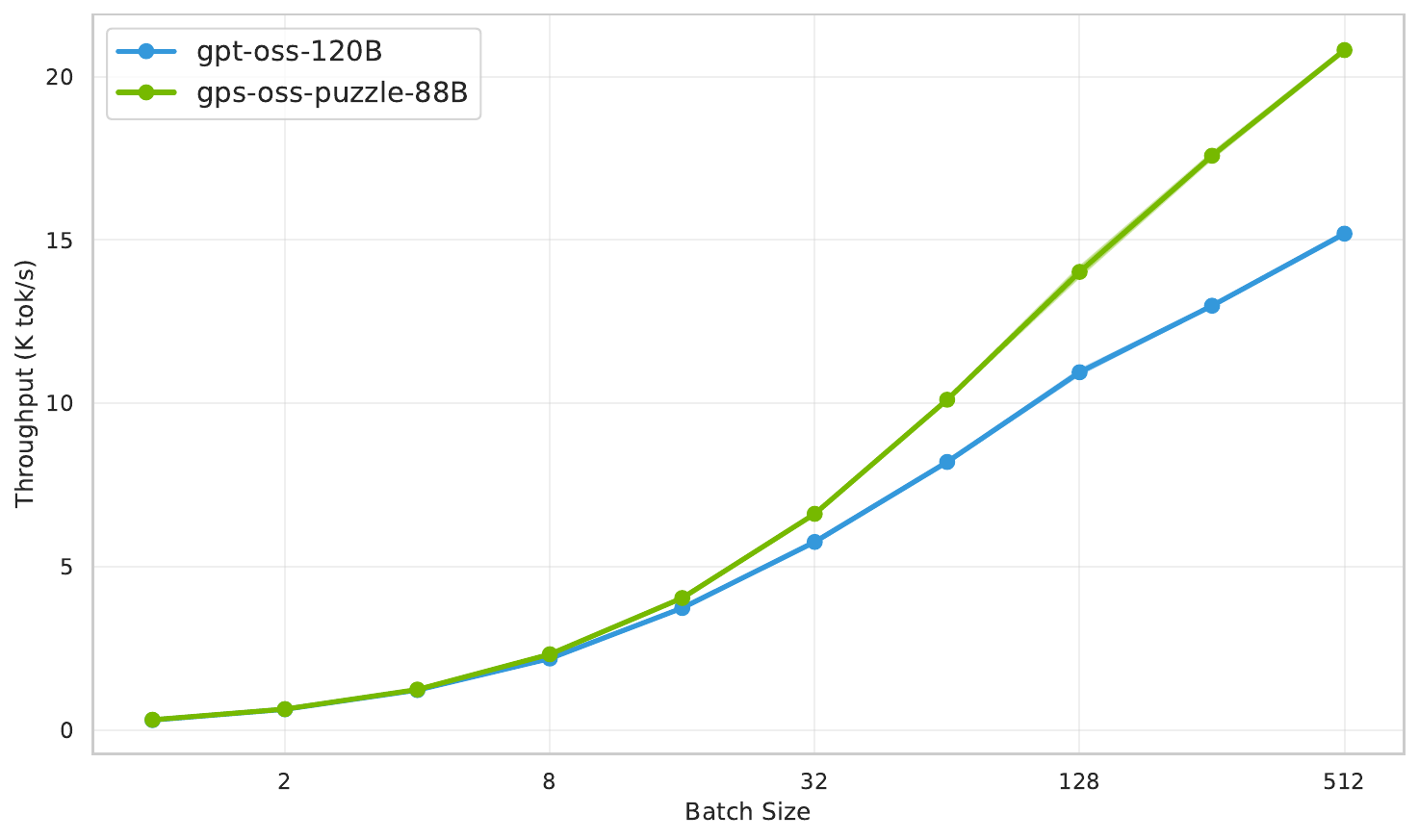}
\caption{B200 throughput scaling with batch size (64K/64K).}
\label{fig:throughput_b200}
\end{figure}

Figure~\ref{fig:throughput_b200} shows throughput scaling with batch size for the 64K/64K scenario.
Across the sweep, the gpt-oss-puzzle-88B consistently outperforms the parent model, with gains increasing in more KV-cache–dominated regimes.

\FloatBarrier

\section{Evaluation Benchmarks Parameters}\label{sec:eval_parameters}

\begin{table}[h]
    \centering
    \caption{Parameters Configuration for Accuracy Benchmarks}
    \label{tab:benchmarks}
    \begin{tabular}{lcccccc}
        \toprule
        \textbf{Benchmark} & \textbf{Temp} & \textbf{Top\_k} & \textbf{Top\_p} & \textbf{Min\_p} & \textbf{Tokens} & \textbf{Reasoning Effort} \\
        \midrule
        *all* & 0.6 & -1 & 0.95 & 0.0 & 128,000 & high/medium/low \\
        \bottomrule
    \end{tabular}
\end{table}

All performance and accuracy measurements were conducted using \texttt{vLLM v0.11.2}; please note that results may vary with different versions.

\FloatBarrier

\section{Ablation Study: Attention Scoring}
\label{ablation_scoring}

\begin{figure}[t]
  \centering
  \includegraphics[width=\textwidth]{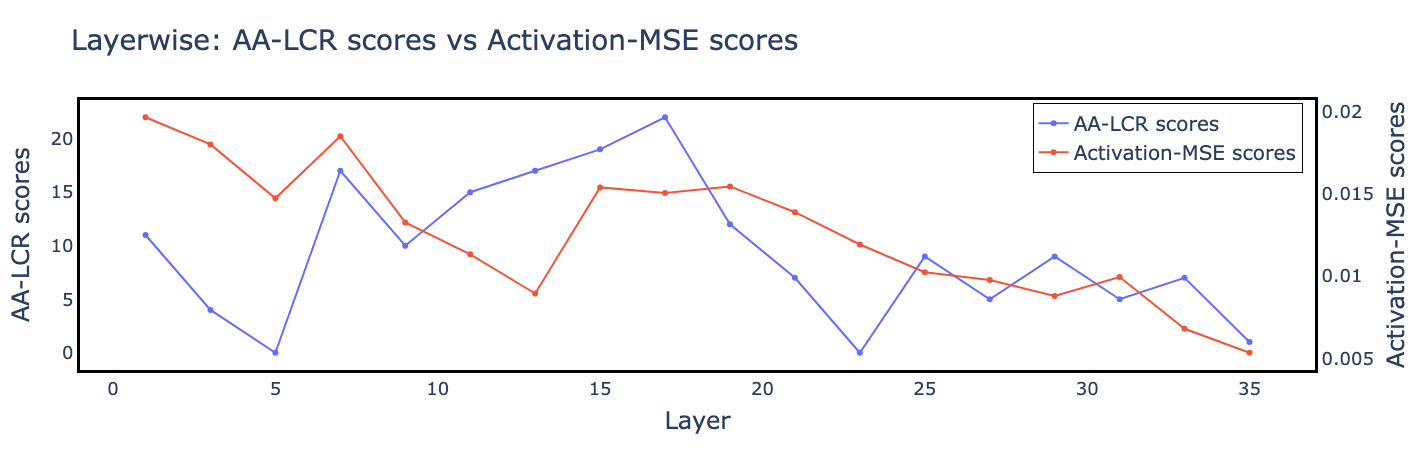}
\caption{Layerwise AA-LCR and activation-MSE replace-1 scores. The AA-LCR score is computed as $\mathrm{AALCR}(\text{parent}) - \mathrm{AALCR}(\text{replace-1})$, and the activation score is the replace-1 activation MSE between the parent and the modified model. Higher values in either metric indicate a larger gap from the parent and therefore a more irreplaceable layer under window-attention replacement.}
\label{fig:aalcr_vs_activation_mse_layerwise}
\end{figure}

\paragraph{Layerwise agreement between AALCR- and activation-based replace-1 scores.}
Figure~\ref{fig:aalcr_vs_activation_mse_layerwise} compares, for each layer, two replace-1 block-quality signals: an AALCR-based gap and an activation-based MSE gap. Specifically, the AALCR score is computed as $\mathrm{AALCR}(\text{parent}) - \mathrm{AALCR}(\text{replace-1})$, while the activation score is the replace-1 MSE between the parent and modified model activations. In both cases, larger values indicate a larger deviation from the parent, suggesting that the corresponding layer is more irreplaceable by window attention. Overall, the two signals largely agree on the relative importance of layers, but they diverge in some cases (e.g., around the intermediate layers), highlighting that representational similarity and task-level degradation can emphasize different sensitivities.

\paragraph{AALCR-scoring and Activation based scoring resulting architecture performance }
In this appendix we compare two block-quality scoring signals used to guide our attention-structure search: regular Puzzle replace-1 scoring versus AALCR-based scoring. The goal in both cases is to assign each candidate window-attention block (i.e., converting one full-attention layer to window attention of size $W$) a scalar quality score, which Puzzle then uses to decide which layers to convert when synthesizing an improved architecture.

In Puzzle replace-1 scoring, for a given layer, we replace the original block (in our case full-attention) with the candidate block (in our case window-attention with window size $W$) while keeping all other layers unchanged. We then run a fixed validation set and measure how closely the modified model matches the parent's internal behavior, using the discrepancy between the last hidden states of the modified model and the parent (e.g., an MSE-style distance). A candidate block is considered higher quality if this single-block replacement perturbs the parent's representations less.

In AALCR-based scoring, we again perform a single-block replacement, but we score block quality by the functional impact on downstream behavior: we compute the drop in AALCR caused by replacing that block, and use this performance degradation as the block-quality signal (smaller drop implies a better block). Once block-quality scores are computed (by either method), we run Puzzle and solve for an improved attention architecture by selecting which layers to convert to window attention under the search constraints. Table~\ref{tab:reg_vs_aalcr_by_cfg_compact_boldbest} summarizes the resulting model-level outcomes when using each scoring signal, enabling a direct comparison of regular replace-1 scoring versus AALCR-based scoring across configurations.

\begin{table}[t]
\centering
\small
\setlength{\tabcolsep}{4pt}
\resizebox{\textwidth}{!}{%
\begin{tabular}{cc|cc cc cc cc cc cc}
\toprule
\textbf{FullAttn} & \textbf{Window}
& \multicolumn{2}{c}{\textbf{MMLU-Pro}}
& \multicolumn{2}{c}{\textbf{GPQA}}
& \multicolumn{2}{c}{\textbf{AIME25}}
& \multicolumn{2}{c}{\textbf{SciCode}}
& \multicolumn{2}{c}{\textbf{AALCR}} \\
\cmidrule(lr){3-4}\cmidrule(lr){5-6}\cmidrule(lr){7-8}\cmidrule(lr){9-10}\cmidrule(lr){11-12}
& & \textbf{Reg} & \textbf{AALCR}
& \textbf{Reg} & \textbf{AALCR}
& \textbf{Reg} & \textbf{AALCR}
& \textbf{Reg} & \textbf{AALCR}
& \textbf{Reg} & \textbf{AALCR} \\
\midrule
9  & 4096
& 76.82 & \textbf{76.89}
& 64.02 & \textbf{64.84}
& 70.63 & \textbf{77.29}
& \textbf{40.24} & 39.35
& 7.00  & \textbf{8.00} \\
9  & 8192
& 78.32 & \textbf{78.92}
& 68.69 & \textbf{73.42}
& 79.17 & \textbf{88.12}
& \textbf{41.12} & 40.24
& 8.00  & \textbf{14.00} \\
12 & 4096
& 78.50 & \textbf{79.55}
& \textbf{74.56} & 74.12
& \textbf{87.78} & 86.88
& 39.64 & \textbf{40.68}
& 15.00 & \textbf{24.00} \\
12 & 8192
& 79.26 & \textbf{79.40}
& 73.67 & \textbf{74.43}
& 88.33 & \textbf{90.00}
& \textbf{40.53} & 40.09
& 17.00 & \textbf{37.00} \\
\bottomrule
\end{tabular}%
}
\caption{Regular scoring vs.\ AALCR-based scoring for selecting window-attention layers. ``FullAttn'' denotes the number of full-attention layers remaining in the final architecture (the parent model has 18 full-attention layers; 12 corresponds to a 33\% reduction and 9 to a 50\% reduction). ``Window'' is the candidate block window size ($4$K or $8$K). ``Reg'' refers to the standard Puzzle replace-1 scoring signal. For each configuration and metric, the better score between Reg and AALCR is bolded.}
\label{tab:reg_vs_aalcr_by_cfg_compact_boldbest}
\end{table}

\section{Raw Results}
\label{app:frontier_raw_results}

This appendix reports the raw numbers underlying Figure~\ref{fig:accuracy_speed_frontier} for a 64K/64K serving scenario on (a) an 8$\times$H100 node and (b) a single H100 GPU.
We also include per-benchmark accuracy and average generated-token statistics for the benchmark suite used to compute the suite average.
We include HyperNova-60B~\citep{hypernova60b} as an external compressed derivative of gpt-oss-120B. HyperNova-60B was evaluated only with KV FP16 (as provided), using the Hugging Face checkpoint available as of Feb.\ 11, 2026.

\paragraph{Relative request rate.}
For each model and reasoning effort, \emph{relative request rate} is computed as max token throughput (best serving configuration per model) divided by the average tokens generated per request, and normalized to the gpt-oss-120B (KV BF16, High reasoning effort) baseline in the corresponding hardware setting.

\paragraph{Max token throughput.}
Tables~\ref{tab:frontier_max_throughput_1node} and~\ref{tab:frontier_max_throughput_1gpu} report the max token throughput values used in the \emph{relative request rate} computation.

\begin{table}[htbp]
\centering
\small
\caption{Max token throughput (64K/64K) at the best configuration per model on an 8$\times$H100 node. The gpt-oss-puzzle-88B derivative achieves 1.63$\times$ and 1.62$\times$ higher throughput than the gpt-oss-120B parent under KV BF16 and KV FP8, respectively.}
\label{tab:frontier_max_throughput_1node}
\setlength{\tabcolsep}{6pt}
\begin{tabular}{llr}
\toprule
Model & KV precision & \makecell{Max throughput\\(K tok/s)} \\
\midrule
gpt-oss-120B & KV BF16 & 4.0 \\
gpt-oss-puzzle-88B & KV BF16 & 6.5 \\
HyperNova-60B & KV BF16 & 6.9 \\
gpt-oss-120B & KV FP8 & 5.8 \\
gpt-oss-puzzle-88B & KV FP8 & 9.3 \\
\bottomrule
\end{tabular}
\end{table}

\begin{table}[htbp]
\centering
\small
\caption{Max token throughput (64K/64K) at the best configuration per model on a single H100 GPU. The gpt-oss-puzzle-88B derivative achieves 2.68$\times$ and 2.86$\times$ higher throughput than the gpt-oss-120B parent under KV BF16 and KV FP8, respectively.}
\label{tab:frontier_max_throughput_1gpu}
\setlength{\tabcolsep}{6pt}
\begin{tabular}{llr}
\toprule
Model & KV precision & \makecell{Max throughput\\(K tok/s)} \\
\midrule
gpt-oss-120B & KV BF16 & 0.2 \\
gpt-oss-puzzle-88B & KV BF16 & 0.5 \\
HyperNova-60B & KV BF16 & 0.7 \\
gpt-oss-120B & KV FP8 & 0.3 \\
gpt-oss-puzzle-88B & KV FP8 & 0.8 \\
\bottomrule
\end{tabular}
\end{table}

\paragraph{Raw numbers used in Figure~\ref{fig:accuracy_speed_frontier}.}
Tables~\ref{tab:frontier_raw_1node} and~\ref{tab:frontier_raw_1gpu} report the suite-average accuracy and relative request rate points plotted in Figure~\ref{fig:accuracy_speed_frontier}.

\begin{table}[htbp]
\centering
\small
\caption{Raw numbers for Figure~\ref{fig:accuracy_speed_frontier}(a): 8$\times$H100 node (64K/64K).}
\label{tab:frontier_raw_1node}
\setlength{\tabcolsep}{5pt}
\begin{tabular}{lllrr}
\toprule
Effort & KV precision & Model & \makecell{Average\\accuracy (\%)} & \makecell{Relative\\request rate} \\
\midrule
\multirow{5}{*}{High} & \multirow{3}{*}{KV BF16} & gpt-oss-120B & 59.20 & 1.000 \\
& & gpt-oss-puzzle-88B & 59.44 & 1.490 \\
& & HyperNova-60B & 51.86 & 1.288 \\
& \multirow{2}{*}{KV FP8} & gpt-oss-120B & 58.19 & 1.401 \\
& & gpt-oss-puzzle-88B & 58.67 & 2.077 \\
\midrule
\multirow{5}{*}{Medium} & \multirow{3}{*}{KV BF16} & gpt-oss-120B & 53.66 & 4.280 \\
& & gpt-oss-puzzle-88B & 56.64 & 4.552 \\
& & HyperNova-60B & 46.23 & 6.032 \\
& \multirow{2}{*}{KV FP8} & gpt-oss-120B & 52.89 & 5.991 \\
& & gpt-oss-puzzle-88B & 54.93 & 6.193 \\
\midrule
\multirow{5}{*}{Low} & \multirow{3}{*}{KV BF16} & gpt-oss-120B & 45.41 & 9.455 \\
& & gpt-oss-puzzle-88B & 50.61 & 11.975 \\
& & HyperNova-60B & 38.77 & 13.855 \\
& \multirow{2}{*}{KV FP8} & gpt-oss-120B & 44.71 & 13.270 \\
& & gpt-oss-puzzle-88B & 48.38 & 17.179 \\
\bottomrule
\end{tabular}
\end{table}

\begin{table}[htbp]
\centering
\small
\caption{Raw numbers for Figure~\ref{fig:accuracy_speed_frontier}(b): single H100 GPU (64K/64K).}
\label{tab:frontier_raw_1gpu}
\setlength{\tabcolsep}{5pt}
\begin{tabular}{lllrr}
\toprule
Effort & KV precision & Model & \makecell{Average\\accuracy (\%)} & \makecell{Relative\\request rate} \\
\midrule
\multirow{5}{*}{High} & \multirow{3}{*}{KV BF16} & gpt-oss-120B & 59.20 & 1.000 \\
& & gpt-oss-puzzle-88B & 59.44 & 2.454 \\
& & HyperNova-60B & 51.86 & 2.934 \\
& \multirow{2}{*}{KV FP8} & gpt-oss-120B & 58.19 & 1.624 \\
& & gpt-oss-puzzle-88B & 58.67 & 4.240 \\
\midrule
\multirow{5}{*}{Medium} & \multirow{3}{*}{KV BF16} & gpt-oss-120B & 53.66 & 4.280 \\
& & gpt-oss-puzzle-88B & 56.64 & 7.497 \\
& & HyperNova-60B & 46.23 & 13.741 \\
& \multirow{2}{*}{KV FP8} & gpt-oss-120B & 52.89 & 6.945 \\
& & gpt-oss-puzzle-88B & 54.93 & 12.644 \\
\midrule
\multirow{5}{*}{Low} & \multirow{3}{*}{KV BF16} & gpt-oss-120B & 45.41 & 9.455 \\
& & gpt-oss-puzzle-88B & 50.61 & 19.724 \\
& & HyperNova-60B & 38.77 & 31.565 \\
& \multirow{2}{*}{KV FP8} & gpt-oss-120B & 44.71 & 15.381 \\
& & gpt-oss-puzzle-88B & 48.38 & 35.076 \\
\bottomrule
\end{tabular}
\end{table}

\paragraph{Per-benchmark accuracy and generation length.}
Table~\ref{tab:frontier_accuracy_tokens_fp16} reports per-benchmark accuracy and average generated tokens for the three compared models, matching the benchmark suite used in Figure~\ref{fig:accuracy_speed_frontier}. gpt-oss-120B and gpt-oss-puzzle-88B use KV BF16; HyperNova-60B was evaluated with KV FP16, as provided.

\begin{table}[htbp]
\centering
\small
\caption{Per-benchmark accuracy (\%) and average tokens generated (K) for the three compared models with KV BF16 in Figure~\ref{fig:accuracy_speed_frontier}.}
\label{tab:frontier_accuracy_tokens_fp16}
\setlength{\tabcolsep}{2.5pt}
\resizebox{\textwidth}{!}{%
\begin{tabular}{llcccccccccc}
\toprule
Effort & Model & \makecell{Average\\accuracy} & \makecell{Average\\tokens (K)} & \makecell{MMLU-\\Pro} & \makecell{RULER\\128K} & \makecell{GPQA-\\Diamond} & \makecell{AIME\\25} & \makecell{IFBench} & \makecell{SciCode} & \makecell{AA\\LCR} & \makecell{HLE} \\
\midrule
\multirow{3}{*}{High} & gpt-oss-120B & 59.20 & 12.70 & 80.41 & 50.89 & 77.78 & 91.46 & 64.46 & 41.72 & 48.75 & 18.16 \\
& gpt-oss-puzzle-88B & 59.44 & 13.87 & 79.32 & 59.80 & 75.13 & 92.92 & 67.77 & 40.83 & 42.25 & 17.52 \\
& HyperNova-60B & 51.86 & 17.09 & 72.97 & 37.27 & 73.93 & 88.75 & 57.03 & 38.31 & 30.00 & 16.64 \\
\midrule
\multirow{3}{*}{Medium} & gpt-oss-120B & 53.66 & 2.97 & 78.86 & 55.01 & 71.28 & 76.88 & 56.55 & 41.42 & 39.25 & 10.06 \\
& gpt-oss-puzzle-88B & 56.64 & 4.54 & 78.03 & 66.71 & 69.70 & 86.88 & 65.56 & 39.64 & 36.00 & 10.57 \\
& HyperNova-60B & 46.23 & 3.65 & 71.03 & 37.09 & 67.42 & 76.25 & 50.00 & 34.02 & 25.00 & 8.99 \\
\midrule
\multirow{3}{*}{Low} & gpt-oss-120B & 45.41 & 1.34 & 75.18 & 52.61 & 62.75 & 50.00 & 44.13 & 39.20 & 35.25 & 4.17 \\
& gpt-oss-puzzle-88B & 50.61 & 1.73 & 75.56 & 66.70 & 64.77 & 66.25 & 56.38 & 38.17 & 31.75 & 5.33 \\
& HyperNova-60B & 38.77 & 1.59 & 66.16 & 42.70 & 59.22 & 44.17 & 40.65 & 31.66 & 20.75 & 4.87 \\
\bottomrule
\end{tabular}
}
\end{table}

\FloatBarrier

\end{document}